\documentclass[runningheads,a4paper]{llncs}
\usepackage[llncs]{styfile}

\usepackage{lineno,hyperref}
\modulolinenumbers[5]
\usepackage{amsmath, amssymb}
\usepackage{latexsym}
\usepackage{graphicx}
\usepackage{epstopdf}
\usepackage{subfig}
\usepackage{float}
\usepackage{url}
\usepackage{multirow}
\usepackage{textcomp}
\usepackage{siunitx}

\begin{document}

\mainmatter  % start of an individual contribution
  
\title{Ensemble of Convolutional Neural Networks for Dermoscopic Images Classification}
\titlerunning{Ensemble of CNNs for Dermoscopic Images Classification}

\newcommand*\samethanks[1][\value{footnote}]{\footnotemark[#1]}
\author{Tom\'{a}\v{s} Majtner\thanks{These two authors contributed equally to this work.}\inst{1}, Buda Baji\'{c}\samethanks\inst{2}, Sule Yildirim\inst{3}, Jon Yngve Hardeberg\inst{3}, Joakim Lindblad\inst{4,5} \and Nata\v{s}a Sladoje\inst{4,5}}
\authorrunning{T. Majtner, B. Baji\'{c}, S. Yildirim, J.Y. Hardeberg, J. Lindblad, N. Sladoje}

\institute{Spanish National Centre for Biotechnology (CNB--CSIC), Madrid, Spain
\and Faculty of Technical Sciences, University of Novi Sad, Serbia
\and Faculty of Information Technology and Electrical Engineering, NTNU, Gj\o{}vik, Norway
\and Centre for Image Analysis, Uppsala University, Sweden
\and Mathematical Institute of the Serbian Academy of Sciences and Arts, Belgrade, Serbia \\
%% E-mails are optional!
%\email{buda.bajic@uns.ac.rs, joakim@cb.uu.se, sladoje@uns.ac.rs}
}
\maketitle 

\begin{abstract}
In this report, we are presenting our automated prediction system for disease classification within dermoscopic images. The proposed solution is based on deep learning, where we employed transfer learning strategy on VGG16 and GoogLeNet architectures. The key feature of our solution is preprocessing based primarily on image augmentation and colour normalization. The solution was evaluated on Task 3: Lesion Diagnosis of the ISIC 2018: Skin Lesion Analysis Towards Melanoma Detection.
\end{abstract}


%%%%%%%%%%%%%%%%%%%%%%%%%%%%%%%%%%%%%%%%%%%%%%
%  Introduction
%%%%%%%%%%%%%%%%%%%%%%%%%%%%%%%%%%%%%%%%%%%%%%
\section{Introduction}
Malignant melanoma, which is commonly known as melanoma, is the most dangerous type of skin cancer \cite{mermelstein1992changing}. It is developed in melanocytes, which are cells producing melanin and and they are giving skin its colour. The risk of melanoma has increasing trend, especially for women under 40. The positive fact is that this form of cancer can be treated successfully with high rate, when it is detected and recognized early. However, the examination of suspicious skin lesions is time-consuming and it requires the knowledge of an expert. It is therefore natural that much research is put to automation of the melanoma recognition process.

The International Skin Imaging Collaboration (ISIC) is an international effort to improve melanoma diagnosis, sponsored by the International Society for Digital Imaging of the Skin (ISDIS). The ISIC Archive contains the largest publicly available collection of quality controlled dermoscopic images of skin lesions. In 2018, ISIC organized already third grand challenge focused on early melanoma detection. This challenge was divided into three separate tasks, where the third tasks was disease classification. 

\begin{figure}[t]
  \centering
  \includegraphics[width=0.79\linewidth]{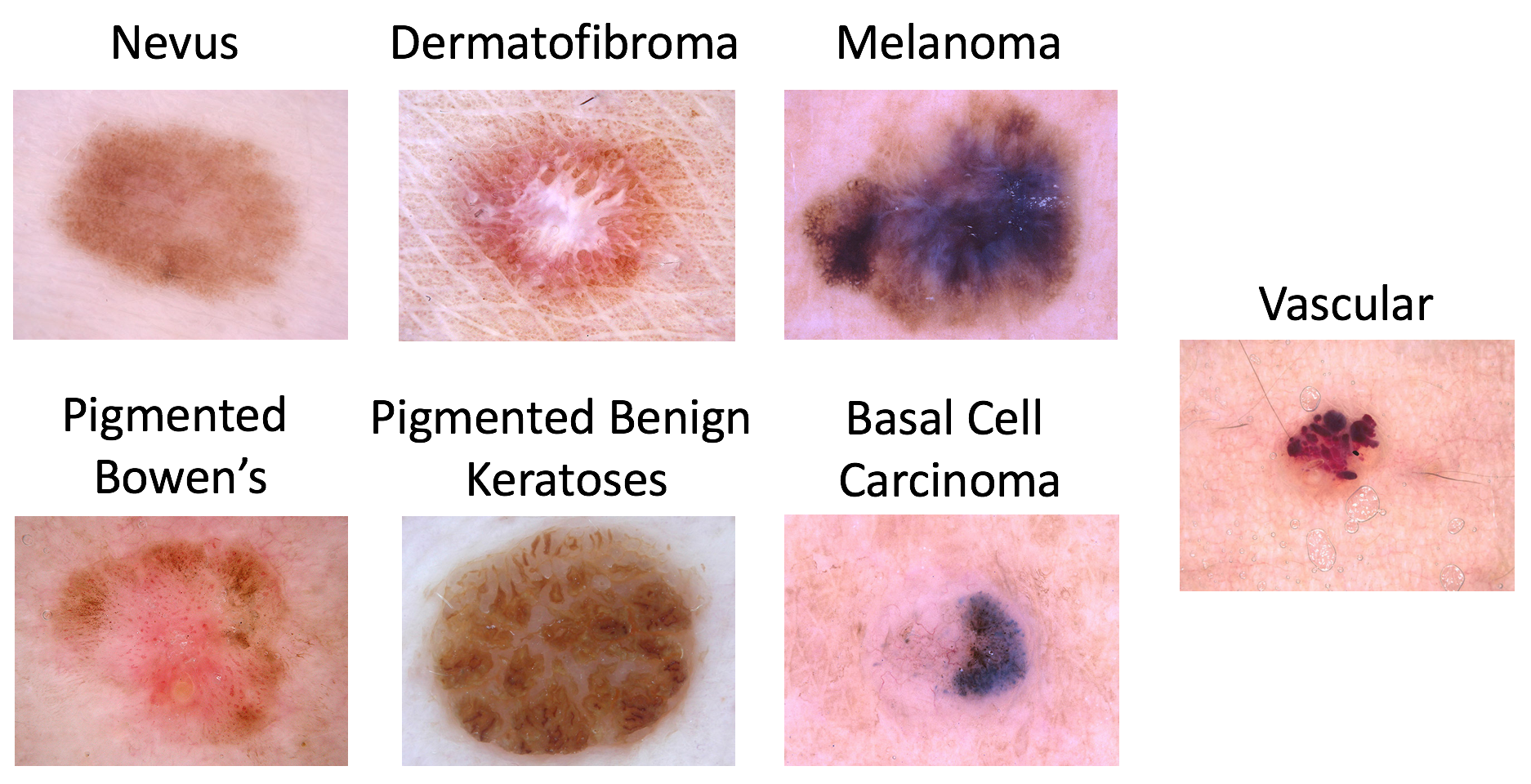}
 \caption{Illustration of disease categories recognized in this study. Image source: https://challenge2018.isic-archive.com/task3/}
 \label{images}
\end{figure}

We are presenting here our framework for automated predictions of disease classification within dermoscopic images. Data for this task was extracted from the "ISIC 2018: Skin Lesion Analysis Towards Melanoma Detection" (ISIC 2018) grand challenge datasets \cite{codella2018skin,tschandl2018ham10000}. The disease categories recognized in this challenge include: Melanoma (MEL), Melanocytic nevus (NV), Basal cell carcinoma (BCC), Actinic keratosis / Bowen's disease (AKIEC), Benign keratosis (BKL), Dermatofibroma (DF), and Vascular lesion (VASC). See Fig.~\ref{images} for illustration.

%%%%%%%%%%%%%%%%%%%%%%%%%%%%%%%%%%%%%%%%%%%%%%
%  Proposed Method  -- Balancing Training Set
%%%%%%%%%%%%%%%%%%%%%%%%%%%%%%%%%%%%%%%%%%%%%%
\section{Proposed Method}
\textit{Balancing Training Set}: The official training set for ISIC 2018 Task 3 competition consists of 10,015 images. As can be seen from the Table~\ref{data_division}, the number of samples in different categories is strongly unbalanced. Moreover, in some particular classes, there is also very small number of images, which could result in inefficient training of automated convolutional neural network (CNN). Therefore, with the aim to increase the number of training images, and at the same time to increase the robustness of our system, we performed data augmentation by horizontal flipping of training samples. This operation double the number of training images in each class. 

The class balancing was achieved by sample rotation. Here, we considered the fact that position of skin lesion during acquisition process is arbitrary and the training data should reflect that property. Rotation was performed around the image center by various rotation factors which are specified in Table~\ref{data_division}. Rotation factor $i$ implies that each image is rotated $i-1$ times. This means that for $i=10$, each images is rotated by angle 36\textdegree$\cdot j$, where $j \in \{1, 2, \dots, 9 \}$.

Presented augmentation leads to more balanced image classes and it also increases rotation invariance of the CNN. Bicubic interpolation is used when the angle was not a multiple of 90\textdegree. As was shown by Goodfellow et al. \cite{goodfellow2016deep}, such data augmentation by flipping and rotation compensates for variations between the training and the test sets and has a positive impact on the CNN performance. Total number of training images after balancing step is 96,274. 
 
\begin{table}
\caption{Total number of training images before and after augmentation.}
\label{data_division}
\centering
	\setlength\tabcolsep{3.5pt}
    \begin{tabular}{c | c c c c c c c}
    \hline
         & AKIEC & BCC & BKL & DF & MEL & NV & VASC \\
         \hline
         Training data & 327 & 514 & 1,099 & 115 & 1,113 & 6,705 & 142 \\
         \hline
         After flipping & 654 & 1,028 & 2,198 & 230 & 2,226 & 13,410 & 284 \\
         \hline
         Rotation factor & 17 & 28 & 60 & 6 & 62 & 0 & 7 \\
         After rotation & 14,388 & 13,364 & 13,188 & 13,800 & 13,356 & 13,410 & 14,768 \\
         \hline
    \end{tabular}
\end{table}

\begin{figure}
	\centering
	\setlength\tabcolsep{1pt}
		\begin{tabular}{cccccc}
		    \vspace{-0.4cm}
            \subfloat{\includegraphics[height=1.3cm]{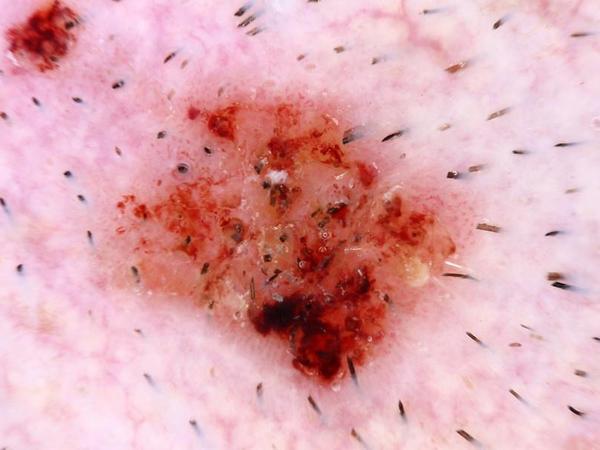}} &
			\subfloat{\includegraphics[height=1.3cm]{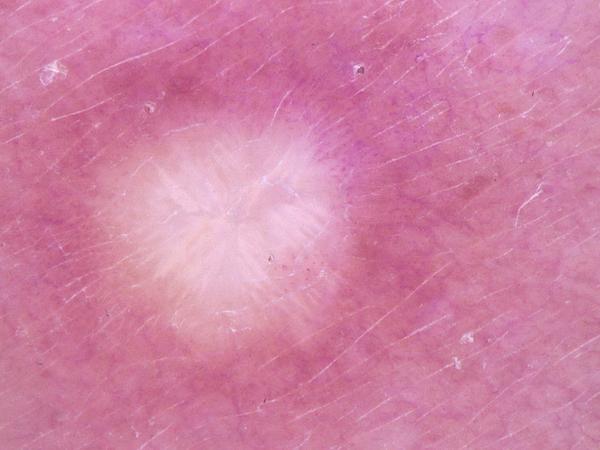}} &
			\subfloat{\includegraphics[height=1.3cm]{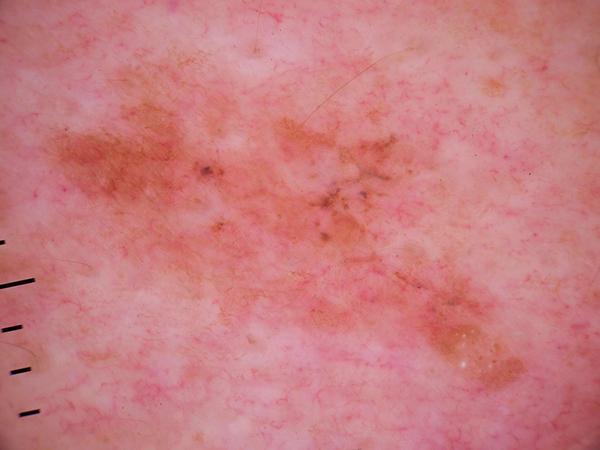}} &
			\subfloat{\includegraphics[height=1.3cm]{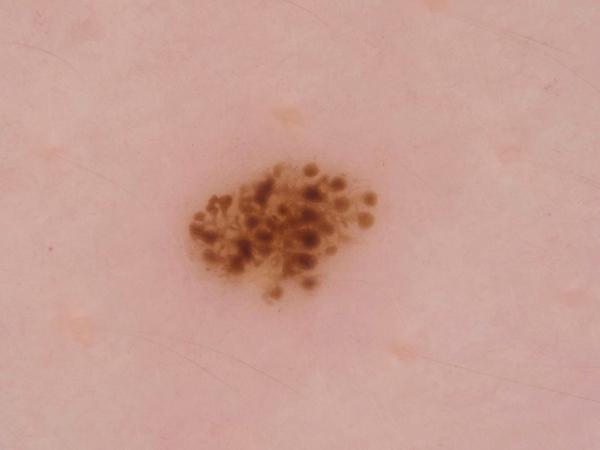}} &
			\subfloat{\includegraphics[height=1.3cm]{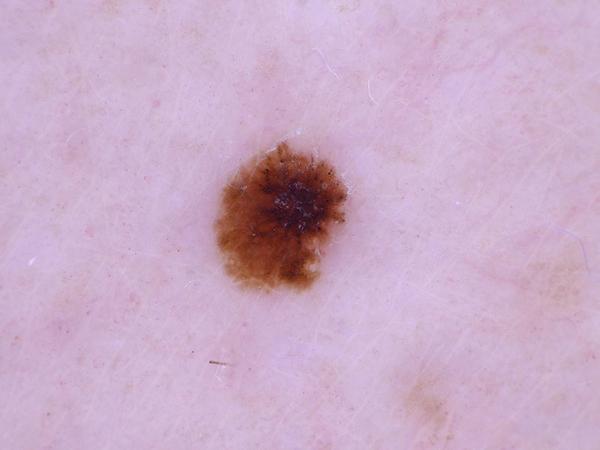}} &
			\subfloat{\includegraphics[height=1.3cm]{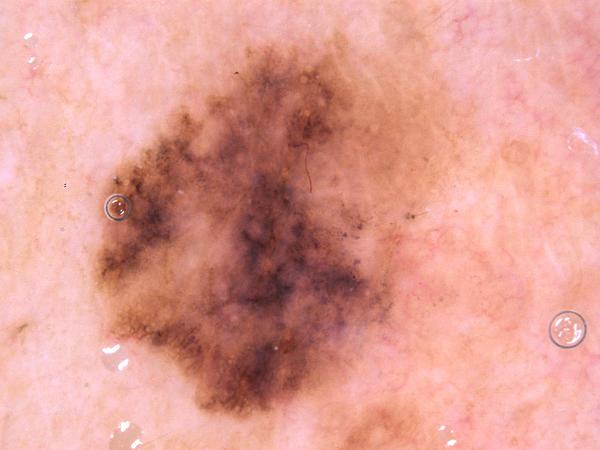}}\\
			\vspace{-0.4cm}
            \subfloat{\includegraphics[height=1.3cm]{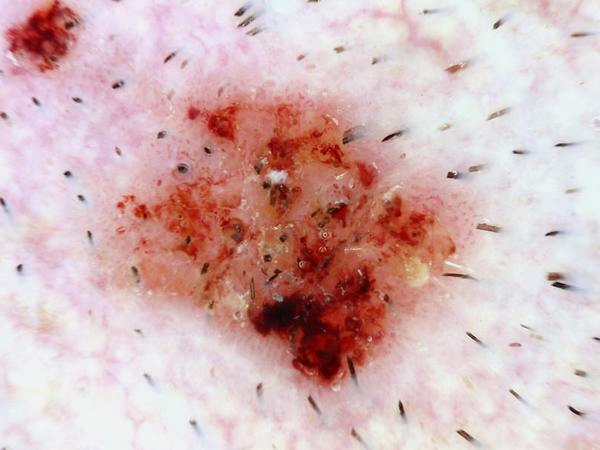}} &
			\subfloat{\includegraphics[height=1.3cm]{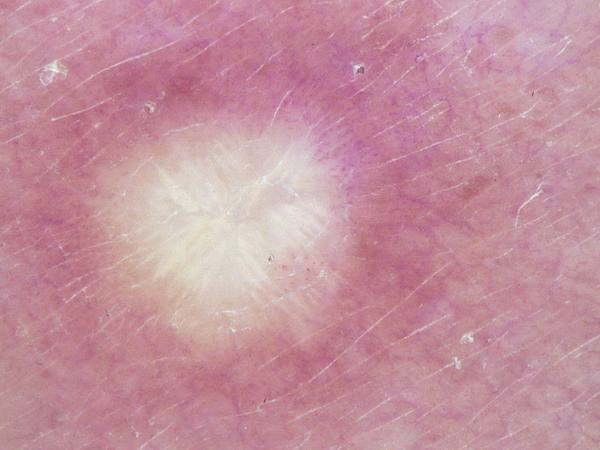}} &
			\subfloat{\includegraphics[height=1.3cm]{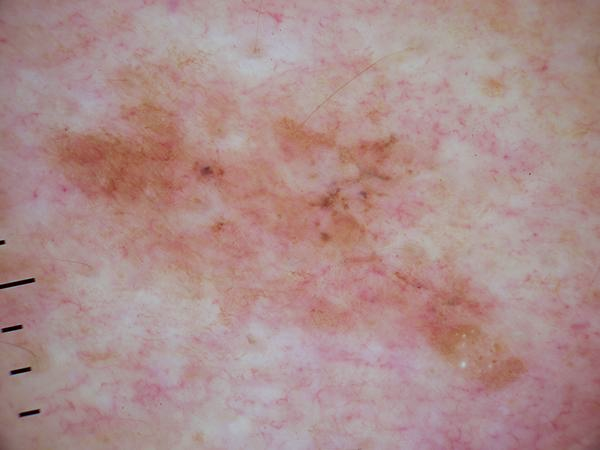}} &
			\subfloat{\includegraphics[height=1.3cm]{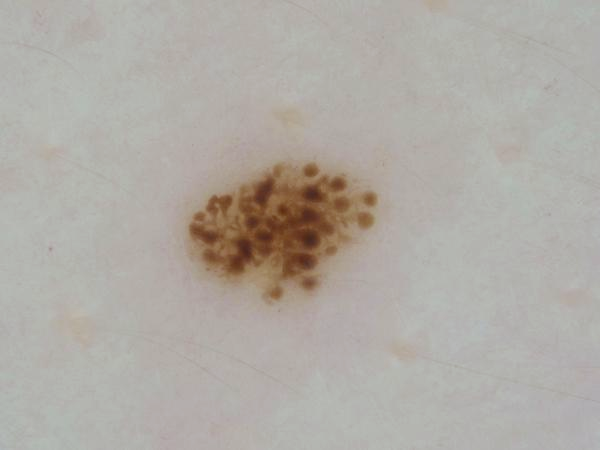}} &
			\subfloat{\includegraphics[height=1.3cm]{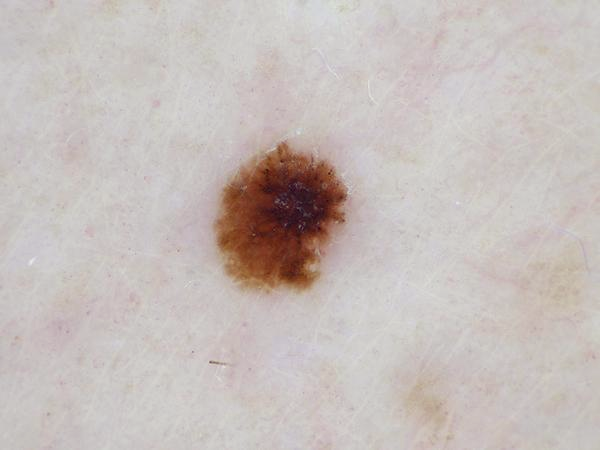}} &
			\subfloat{\includegraphics[height=1.3cm]{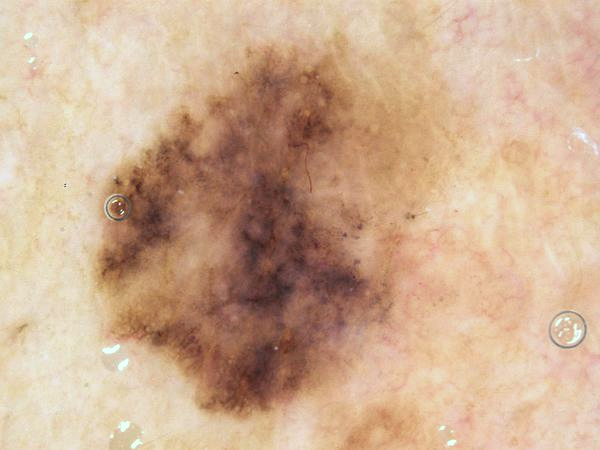}}
			\vspace{0.2cm}
		\end{tabular}
	\caption{Examples of image samples after colour normalization. First row corresponds to original images and second row to max-RGB normalization.}	
	\label{colout_constancy}		
\end{figure}

%%%%%%%%%%%%%%%%%%%%%%%%%%%%%%%%%%%%%%%%%%%%%%
%  Proposed Method  -- Colour Normalization
%%%%%%%%%%%%%%%%%%%%%%%%%%%%%%%%%%%%%%%%%%%%%%
\textit{Colour Normalization}: Because the images are from different sources, various lightning conditions change the visual appearance of the skin lesions. Traditional conversion to grayscale images would lead to significant lost of input information, therefore we aimed for colour constancy algorithms. This topic was already covered in number of papers \cite{madooei2016incorporating}. In our work, we tested several methods and decided to use max-RGB \cite{land1977retinex}, which was also used in a recent study \cite{barata2015improving}. Max-RGB method is based on the assumption that the reflectance, which is achieved for each of the three channels, is equal \cite{van2007edge}. The illustration of this method is presented in Fig.~\ref{colout_constancy}.

%%%%%%%%%%%%%%%%%%%%%%%%%%%%%%%%%%%%%%%%%%%%%%
%  Proposed Method  -- CNNs 
%%%%%%%%%%%%%%%%%%%%%%%%%%%%%%%%%%%%%%%%%%%%%%
\textit{Transfer Learning}: Our classification is based on fine-tunning the VGG16 network \cite{simonyan2014very} and the GoogLeNet network \cite{szegedy2015going}. As it was observed in number of previous works \cite{tajbakhsh2016convolutional}, training a deep convolutional neural network from scratch is difficult because it requires a large amount of labeled training data which in case of medical applications is rarely available. A promising alternative is to fine-tune of a CNN that has been pre-trained using, for instance, a large set of labeled natural images such as ImageNet dataset. 

The usage of transfer learning for skin lesion classification in this work was additionally motivated by recent skin cancer classification study \cite{esteva2017dermatologist}. It was performed by retraining the weights of the last three layers, where our version has seven neurons in its output layer. For training purposes, we used colour normalized augmented training dataset. We utilize stochastic gradient descent (SGD) for optimization, with a momentum factor of 0.9, and $\ell_2$ regularization level of $10^{-4}$. The learning rate of 0.0001 was used in a mini-batch scheme of 8 images and employing 50 epochs for retraining. Our solution was implemented using \textsc{Matlab} R2018a.

%%%%%%%%%%%%%%%%%%%%%%%%%%%%%%%%%%%%%%%%%%%%%%
%  Proposed Method - Ensembling
%%%%%%%%%%%%%%%%%%%%%%%%%%%%%%%%%%%%%%%%%%%%%%
\textit{Ensembling}: As a last step, we employed ensemble of both our VGG16 and GoogLeNet networks fine-tuned with preprocessed images. The ensemble typically leads to higher performance, when compared with results of a single network \cite{kumar2017ensemble}. In our case, the ensemble weights were selected manually to be 0.5 for both networks.

%%%%%%%%%%%%%%%%%%%%%%%%%%%%%%%%%%%%%%%%%%%%%%
%  Results
%%%%%%%%%%%%%%%%%%%%%%%%%%%%%%%%%%%%%%%%%%%%%%
\section{Results}
The proposed solution was evaluated using the official validation set for ISIC 2018 Task 3 competition. This set consists of 193 images and it was evaluated automatically via submission webpage. The achieved balanced accuracy was 0.801 for the VGG16 architecture, 0.797 for the GoogLeNet architecture, and 0.815 for their ensemble. 

\bibliographystyle{splncs03} 
\bibliography{references}

\end{document}